\documentclass{article}
\usepackage{spconf,amsmath,graphicx}
\usepackage{todonotes}
\usepackage{float}
\usepackage{amssymb,amsmath,comment, graphicx,amsmath,mathrsfs, array, hyperref, multirow}
\usepackage[caption=false]{subfig}
\usepackage{booktabs}
\usepackage{diagbox}
\usepackage{xcolor}
\usepackage{tabularx,booktabs}

\title{Hierarchical Transformers for Long Document Classification}

\name{Raghavendra Pappagari$^{1}$, Piotr \.Zelasko$^{2}$,
        Jes\'us Villalba$^{1}$, Yishay Carmiel$^{2}$, 
        and~Najim Dehak$^{1}$}
\address{$^{1}$Center for Language and Speech Processing,
        Johns Hopkins University, Baltimore, MD \\
        $^{2}$Avaya Conversational Intelligence \\
\tt \{rpappag1,jvillal7,ndehak3\}@jhu.edu \\
\tt    petezor@gmail.com, ycarmiel@avaya.com} 
\begin{document}
%
\maketitle
\begin{abstract}
BERT, which stands for Bidirectional Encoder Representations from Transformers, is a recently introduced language representation model based upon the transfer learning paradigm. 
We extend its fine-tuning procedure to address one of its major limitations - applicability to inputs longer than a few hundred words, such as transcripts of human call conversations. 
Our method is conceptually simple. We segment the input into smaller chunks and feed each of them into the base model. 
Then, we propagate each output through a single recurrent layer, or another transformer, followed by a softmax activation. 
We obtain the final classification decision after the last segment has been consumed. 
We show that both BERT extensions are quick to fine-tune and converge after as little as 1 epoch of training on a small, domain-specific data set. 
We successfully apply them in three different tasks involving customer call satisfaction prediction and topic classification, and obtain a significant improvement over the baseline models in two of them.
\end{abstract}
\begin{keywords}
Transformer, BERT, Recurrent Neural Networks,Topic Identification, Customer Satisfaction Prediction
\end{keywords}
\section{Introduction}
\label{sec:intro}

%


Bidirectional Encoder Representations from Transformers (BERT) is a novel Transformer~\cite{vaswani2017attention} model, which recently achieved state-of-the-art performance in several language understanding tasks, such as question answering, natural language inference, semantic similarity, sentiment analysis, and others~\cite{devlin2018bert}. 
While well-suited to dealing with relatively short sequences, Transformers suffer from a major issue that hinders their applicability in classification of long sequences, i.e. they are able to consume only a limited context of symbols as their input~\cite{dai2019transformer}.

There are several natural language (NLP) processing tasks that involve such long sequences. 
Of particular interest are topic identification of spoken conversations~\cite{TopicID_Fisher_MCE,TopicID_SMM,pappagari2018joint} and call center customer satisfaction prediction~\cite{Turn_taking_feats,CSAT_CNN,CSAT_text_SVM_IBM, automotive_industry}. 
Call center conversations, while usually quite short and to the point, often involve agents trying to solve very complex issues that the customers experience, resulting in some calls taking even an hour or more.
For speech analytics purposes, these calls are typically transcribed using an automatic speech recognition (ASR) system, and processed in textual representations further down the NLP pipeline.
These transcripts sometimes exceed the length of 5000 words. 
Furthermore, temporal information might play an important role in tasks like CSAT.
For example, a customer may be angry at the beginning of the call, but after her issue is resolved, she would be very satisfied with the way it was handled. 
Therefore, simple bag of words models, or any model that does not include temporal dependencies between the inputs, may not be well-suited to handle this category of tasks.
This motivates us to employ model such as BERT in this task.

In this paper, we propose a method that builds upon BERT's architecture. 
We split the input text sequence into shorter segments in order to obtain a representation for each of them using BERT. 
Then, we use either a recurrent LSTM~\cite{hochreiter1997long} network, or another Transformer, to perform the actual classification. 
We call these techniques \textbf{R}ecurrence \textbf{o}ver \textbf{BERT} (\textbf{RoBERT}) and \textbf{T}ransformer \textbf{o}ver \textbf{BERT} (\textbf{ToBERT}). 
Given that these models introduce a hierarchy of representations (segment-wise and document-wise), we refer to them as Hierarchical Transformers. 
To the best of our knowledge, no attempt has been done before to use the Transformer architecture for classification of such long sequences.


Our novel contributions are:
\begin{itemize}
    \item Two extensions - RoBERT and ToBERT - to the BERT model, which enable its application in classification of long texts by performing segmentation and using another layer on top of the segment representations.
    \item State-of-the-art results on the Fisher topic classification task.
    \item Significant improvement on the CSAT prediction task over the MS-CNN model.
\end{itemize}



\section{Related work}
\label{sec:related_work}

Several dimensionality reduction algorithms such as RBM, autoencoders, subspace multinomial models (SMM) are used to obtain a low dimensional representation of documents from a simple BOW representation and then classify it using a simple linear classifiers~\cite{TopicID_Replicated_Softmax, TopicID_RBM, TopicID_Kate, TopicID_SMM}. 
In~\cite{yang2016hierarchical} hierarchical attention networks are used for document classification. They evaluate their model on several datasets with average number of words around 150. 
Character-level CNN are explored in~\cite{zhang2015character} but it is prohibitive for very long documents.
In~\cite{liu2018long}, dataset collected from arXiv papers is used for classification. For classification, they sample random blocks of words and use them together for classification instead of using full document which may work well as arXiv papers are usually coherent and well written on a well defined topic. Their method may not work well on spoken conversations as random block of words usually do not represent topic of full conversation.

Several researchers addressed the problem of predicting customer satisfaction~\cite{Turn_taking_feats,CSAT_CNN,CSAT_text_SVM_IBM, automotive_industry}. In most of these works, logistic regression, SVM, CNN  are applied on different kinds of representations.

In~\cite{adhikari2019docbert}, authors use BERT for document classification but the average document length is less than BERT maximum length 512.
TransformerXL~\cite{dai2019transformer} is an extension to the Transformer architecture that allows it to better deal with long inputs for the language modelling task. It relies on the auto-regressive property of the model, which is not the case in our tasks.

\section{Method}
\label{sec:format}

\subsection{BERT}

\begin{figure}
    \centering
    \includegraphics[scale=0.5]{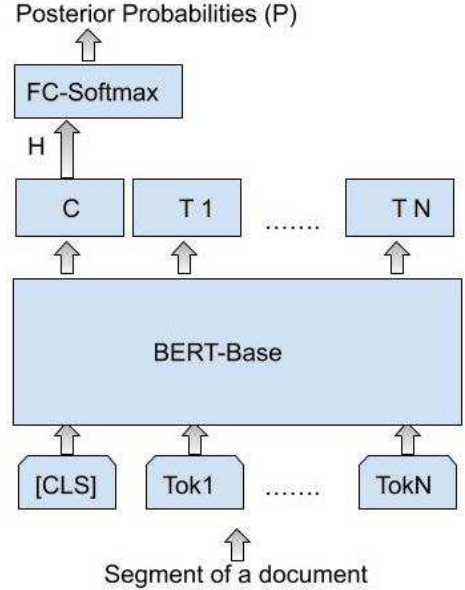}
    \caption{BERT model for classification. H denotes BERT segment representations from last transformer block, P denotes segment posterior probabilities. Figure inspired from~\cite{devlin2018bert}}
    \label{fig:bert_classif}
\end{figure}

Because our work builds heavily upon BERT, we provide a brief summary of its features. 
BERT is built upon the Transformer architecture~\cite{vaswani2017attention}, which uses self-attention, feed-forward layers, residual connections and layer normalization as the main building blocks. 
It has two pre-training objectives:
\begin{itemize}
    \item Masked language modelling - some of the words in a sentence are being masked and the model has to predict them based on the context (note the difference from the typical autoregressive language model training objective);
    \item Next sentence prediction - given two input sequences, decide whether the second one is the next sentence or not.
\end{itemize}
BERT has been shown to beat the state-of-the-art performance on 11 tasks with no modifications to the model architecture, besides adding a task-specific output layer~\cite{devlin2018bert}.
We follow same procedure suggested in~\cite{devlin2018bert} for our tasks.
Fig.~\ref{fig:bert_classif} shows the BERT model for classification. 
We obtain two kinds of representation from BERT: pooled output from last transformer block, denoted by H, and posterior probabilities, denoted by P.
There are two variants of BERT - BERT-Base and BERT-Large. 
In this work we are using BERT-Base for faster training and experimentation, however, our methods are applicable to BERT-Large as well. 
BERT-Base and BERT-Large are different in model parameters such as number of transformer blocks, number of self-attention heads. 
Total number of parameters in BERT-Base are 110M and 340M in BERT-Large.

BERT suffers from major limitations in terms of handling long sequences. 
Firstly, the self-attention layer has a quadratic complexity $O(n^2)$ in terms of the sequence length $n$~\cite{vaswani2017attention}.
Secondly, BERT uses a learned positional embeddings scheme~\cite{devlin2018bert}, which means that it won't likely be able to generalize to positions beyond those seen in the training data.

To investigate the effect of fine-tuning BERT on task performance, we use either the pre-trained BERT weights\footnote{Available at https://github.com/huggingface/pytorch-pretrained-BERT}, or the weights from a BERT fine-tuned on the task-specific dataset on a segment-level (i.e. we preserve the original label but fine-tune on each segment separately instead of on the whole text sequence).
We compare these results to using the fine-tuned segment-level BERT predictions directly as inputs to the next layer.

\subsection{Recurrence over BERT}

Given that BERT is limited to a particular input length, we split the input sequence into segments of a fixed size with overlap.
For each of these segments, we obtain H or P from BERT model.
We then stack these segment-level representations into a sequence, which serves as input to a small (100-dimensional) LSTM layer.
Its output serves as a \emph{document embedding}.
Finally, we use two fully connected layers with ReLU (30-dimensional) and softmax (the same dimensionality as the number of classes) activations to obtain the final predictions.

With this approach, we overcome BERT's computational complexity, reducing it to $O(n/k * k^2) = O(nk)$ for RoBERT, with $k$ denoting the segment size (the LSTM component has negligible linear complexity $O(k)$).
The positional embeddings are also no longer an issue.

\subsection{Transformer over BERT}

Given that Transformers' edge over recurrent networks is their ability to effectively capture long distance relationships between words in a sequence~\cite{vaswani2017attention}, we experiment with replacing the LSTM recurrent layer in favor of a small Transformer model (2 layers of transformer  building block containing self-attention, fully connected, etc.).
To investigate if preserving the information about the input sequence order is important, we also build a variant of ToBERT which learns positional embeddings at the segment-level representations (but is limited to sequences of length seen during the training).

ToBERT's computational complexity $O(\frac{n^2}{k^2})$ is asymptotically inferior to RoBERT, as the top-level Transformer model again suffers from quadratic complexity in the number of segments. However, in practice this number is much smaller than the input sequence length (${\frac{n}{k}} << n$), so we haven't observed performance or memory issues with our datasets.

\section{Experiments}
\label{sec:pagestyle}

We evaluated our models on 3 different datasets: 
\begin{itemize}
    \item CSAT dataset for CSAT prediction, consisting of spoken transcripts (automatic via ASR).
    \item 20 newsgroups for topic identification task, consisting of written text;
    \item Fisher Phase 1 corpus for topic identification task, consisting of spoken transcripts (manual);
\end{itemize}

\subsection{CSAT}
\label{subsec:csat_data}
\begin{figure}
    \centering
    \includegraphics[width=0.45\textwidth, scale=1]{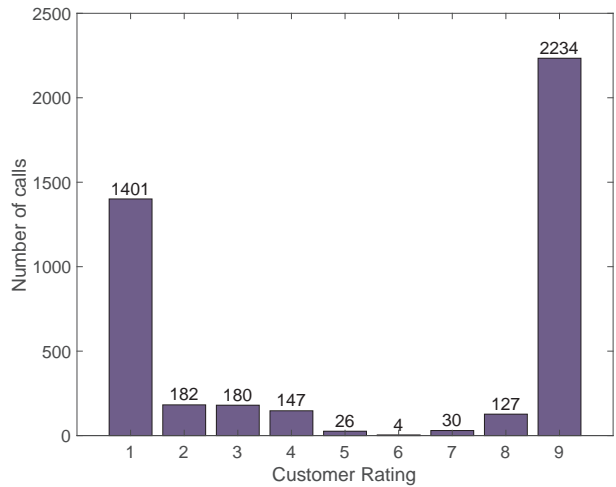}
    \caption{Histogram of customer ratings. Rating 9 corresponds to extremely satisfied and 1 to extremely dissatisfied}
    \label{fig:Hist_FS_rating}
\end{figure}

CSAT dataset consists of US English telephone speech from call centers.
For each call in this dataset, customers participated in that call gave a rating on his experience with agent.
Originally, this dataset has labels rated on a scale 1-9 with 9 being extremely satisfied and 1 being extremely dissatisfied. 
Fig.~\ref{fig:Hist_FS_rating} shows the histogram of ratings for our dataset.
As the distribution is skewed towards extremes, we choose to do binary classification with ratings above 4.5 as satisfied and below 4.5 as dissatisfied.
Quantization of ratings also helped us to create a balanced dataset.
This dataset contains 4331 calls and we split them into 3 sets for our experiments: 2866 calls for training, 362 calls for validation and, finally, 1103 calls for testing.

We obtained the transcripts by employing an ASR system.
The ASR system uses TDNN-LSTM acoustic model trained on Fisher and Switchboard datasets with lattice-free maximum mutual information criterion~\cite{LF_MMI_ASR}.
The word error rates using four-gram language models were 9.2\% and 17.3\% respectively on Switchboard and CallHome portions of Eval2000 dataset\footnote{https://catalog.ldc.upenn.edu/LDC2002T43}.

\subsection{20 newsgroups}
20 newsgroups\footnote{http://qwone.com/~jason/20Newsgroups/} data set is one of the frequently used datasets in the text processing community for text classification and text clustering. 
This data set contains approximately 20,000 English documents from 20 topics to be identified, with 11314 documents for training and 7532 for testing.
In this work, we used only 90\% of documents for training and the remaining 10\% for validation.
For fair comparison with other publications, we used 53160 words vocabulary set available in the datasets website.

\subsection{Fisher}
Fisher Phase 1 US English corpus is often used for automatic speech recognition in speech community.
In this work, we used it for topic identification as in~\cite{TopicID_Fisher_MCE}.
The documents are 10-minute long telephone conversations between two people discussing a given topic.
We used same training and test splits as~\cite{TopicID_Fisher_MCE} in which 1374 and 1372 documents are used for training and testing respectively.
For validation of our model, we used 10\% of training dataset and the remaining 90\% was used for actual model training.
The number of topics in this data set is 40.

\subsection{Dataset Statistics}
\begin{figure}
    \centering
        \includegraphics[width=0.45\textwidth,scale=1]{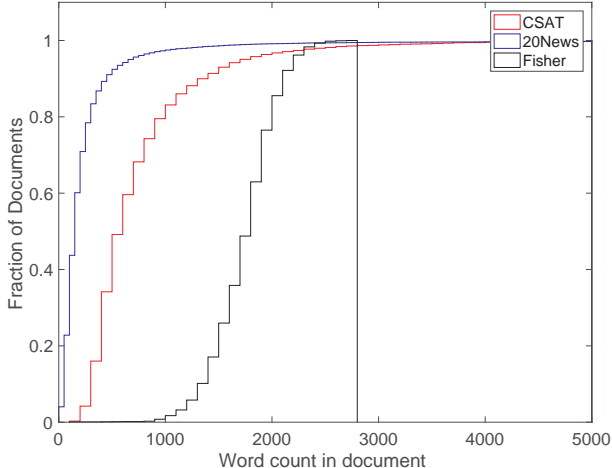}
    \caption{Cumulative distribution of document lengths.}
    \label{fig:length_dist}
\end{figure}

Table~\ref{tab:dataset_stats} shows statistics of our datasets.
It can be observed that average length of Fisher is much higher than 20 newsgroups and CSAT.
Cumulative distribution of document lengths for each dataset is shown in Fig.~\ref{fig:length_dist}.
It can be observed that almost all of the documents in Fisher dataset have length more than 1000 words.
For CSAT, more than 50\% of the documents have length greater than 500 and for 20newsgroups only 10\% of the documents have length greater than 500.
Note that, for CSAT and 20newsgroups, there are few documents with length more than 5000.

\begin{table}
    \centering
    \begin{tabular}{@{}l|cccc@{}}
    \toprule
    Dataset & C & N & AW & L \\
    \midrule
    CSAT & 2 &  4331   &  787 & 10503\\
    20 newsgroups & 20 &  18846 &  266 & 10334\\
    Fisher & 40 & 2746 & 1788 & 2713 \\
    \bottomrule
    \end{tabular}
    \caption{Dataset statistics. C indicates number of Classes, N the Number of documents, AW the Average number of Words per document and L the Longest document length.}
    \label{tab:dataset_stats}
\end{table}

\subsection{Architecture and Training Details}

In this work, we split document into segments of 200 tokens with a shift of 50 tokens to extract features from BERT model.
For RoBERT, LSTM model is trained to minimize cross-entropy loss with \textit{Adam} optimizer~\cite{kingma2014adam}.
The initial learning rate is set to $0.001$ and is reduced by a factor of $0.95$ if validation loss does not decrease for 3-epochs. 
For ToBERT, the Transformer is trained with the default BERT version of \textit{Adam} optimizer~\cite{devlin2018bert} with an initial learning rate of $5e$-$5$.
We report accuracy in all of our experiments.
We chose a model with the best validation accuracy to calculate accuracy on the test set.
To accomodate for non-determinism of some TensorFlow\footnote{We used TensorFlow version 1.14.0.} GPU operations, we report accuracy averaged over 5 runs.


\section{Results}

Table~\ref{tab:pretrain_bert_results} presents results using pre-trained BERT features. 
We extracted features from the pooled output of final transformer block as these were shown to be working well for most of the tasks~\cite{devlin2018bert}.
The features extracted from a pre-trained BERT model without any fine-tuning lead to a sub-par performance.
However, We also notice that ToBERT model exploited the pre-trained BERT features better than RoBERT.
It also converged faster than RoBERT.
Table~\ref{tab:finetune_bert_results} shows results using features extracted after fine-tuning BERT model with our datasets.
Significant improvements can be observed compared to using pre-trained BERT features.
Also, it can be noticed that ToBERT outperforms RoBERT on Fisher and 20newsgroups dataset by 13.63\% and 0.81\% respectively.
On CSAT, ToBERT performs slightly worse than RoBERT but it is not statistically significant as this dataset is small. 


\begin{table}
    \centering
    \begin{tabular}{@{}l|c|c@{}}
    \toprule
     &RoBERT & ToBERT \\
     \midrule
    CSAT  & 71.16 & 74.77    \\
    20newsgroups & 60.75 & 65.04   \\
    Fisher & 38.04 &80.68  \\
    \bottomrule
    \end{tabular}
    \caption{Results using segment representations (H) from a pre-trained BERT (without fine-tuning).}
    \label{tab:pretrain_bert_results}
\end{table}

\begin{table}
    \centering
    \begin{tabular}{@{}l|c|c@{}}
    \toprule
     & RoBERT & ToBERT\\
    \midrule
    CSAT  & 83.65   & 83.48  \\
    20newsgroups &  84.71 & 85.52  \\
    Fisher & 82.28 & 95.48   \\
    \bottomrule
    \end{tabular}
    \caption{Results using segment representations (H) from a fine-tuned BERT.}
    \label{tab:finetune_bert_results}
\end{table}

\begin{table}
    \centering
    \resizebox{\columnwidth}{!}{
    \begin{tabular}{c|c|c|c|c}
    \toprule
        & Most frequent & Average & RoBERT & ToBERT \\
         \midrule
    CSAT  & 81.03   & 82.84& \textbf{83.54}  & 81.48    \\
    20newsgroups  & 84.78 &  84.51  &  84.07 & \textbf{85.47} \\
    Fisher & 88.70 &  88.48 &91.18  & \textbf{94.16}   \\
    \bottomrule
    \end{tabular}}
    \caption{Comparison of models using fine-tuned BERT segment-level predictions (P) instead of segment representations (H). }
    \label{tab:bert_pred_results}
\end{table}

Table~\ref{tab:bert_pred_results} presents results using fine-tuned BERT predictions instead of the pooled output from final transformer block.
For each document, having obtained segment-wise predictions we can obtain final prediction for the whole document in three ways: 
\begin{itemize}
\item Compute the average of all segment-wise predictions and find the most probable class;
\item Find the most frequently predicted class;
\item Train a classification model.
\end{itemize}

\begin{figure}
    \centering
        \includegraphics[width=0.45\textwidth,scale=1]{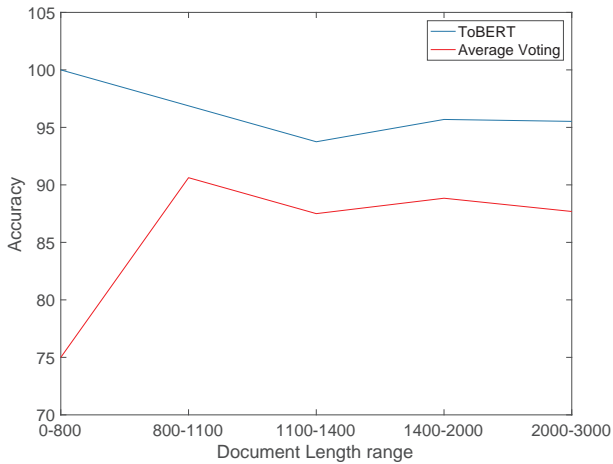}
    \caption{Comparison of average voting and ToBERT for various document length ranges for Fisher dataset.}
    \label{fig:acc_bucketing}
\end{figure}

It can be observed from Table~\ref{tab:bert_pred_results} that a simple averaging operation or taking most frequent predicted class works competitively for CSAT and 20newsgroups but not for the Fisher dataset.
We believe the improvements from using RoBERT or ToBERT, compared to simple averaging or most frequent operations, are proportional to the fraction of long documents in the dataset. 
CSAT and 20newsgroups have (on average) significantly shorter documents than Fisher, as seen in Fig.~\ref{fig:length_dist}.
Also, significant improvements for Fisher could be because of less confident predictions from BERT model as this dataset has 40 classes.
Fig.~\ref{fig:acc_bucketing} presents the comparison of average voting and ToBERT for various document length ranges for Fisher dataset. We used fine-tuned BERT segment-level predictions (P) for this analysis.
It can be observed that ToBERT outperforms average voting in every interval.
To the best of our knowledge, this is a state-of-the-art result reported on the Fisher dataset.

\begin{table}
    \centering
    \begin{tabular}{@{}l|ccc@{}}
    \toprule
    \multirow{2}{*}{}& \multicolumn{2}{c}{Position embeddings}   \\
    \cline{2-3}
     & No & Yes   \\
     \midrule
    CSAT  & 82.84  & \textbf{83.48} \\
    20newsgroups & 85.51 & \textbf{85.52}   \\
    Fisher & 95.84  & \textbf{95.48}  \\
    \bottomrule
    \end{tabular}
    \caption{The effect of including positional embeddings in ToBERT model. Fine-tuned BERT segment representations were used for these results. }
    \label{tab:pos_emb_effect}
\end{table}

Table~\ref{tab:pos_emb_effect} presents the effect of position embeddings on the model performance.
It can be observed that position embeddings did not significantly affect the model performance for Fisher and 20newsgroups, but they helped slightly in CSAT prediction (an absolute improvement of 0.64\% F1-score). 
We think that this is explained by the fact that Fisher and 20newsgroups are topic identification tasks, and the topic does not change much throughout these documents.
However, CSAT may vary during the call, and in some cases a naive assumption that the sequential nature of the transcripts is irrelevant may lead to wrong conclusions.


Table~\ref{tab:comp_literature} compares our results with previous works. 
It can be seen that our model ToBERT outperforms CNN based experiments by significant margin on CSAT and Fisher datasets.
For CSAT dataset, we used multi-scale CNN (MS-CNN) as the baseline, given its strong results on Fisher and 20newsgroups. 
The setup was replicated from~\cite{pappagari2018joint} for comparison.
We also see that our result on 20 newsgroups is 0.6\% worse than the state-of-the-art.



\begin{table}
\centering
\scalebox{1}
{
\renewcommand{\arraystretch}{1.4}
\begin{normalsize}
\begin{tabular}{@{}l|ccc@{}}
\hline
dataset &Model & Accuracy \\
\hline
\multirow{2}{*}{CSAT} & MS-CNN & 79.53 \\
& ToBERT & 83.48 \\ 
& RoBERT & \textbf{83.65}  \\
\hline 
\multirow{3}{*}{20 newsgroups} & SCDV \cite{TopicID_SCDV} & 84.6  \\
&MS-CNN~\cite{pappagari2018joint} & \textbf{86.12}   \\
& ToBERT & 85.52 \\
& RoBERT & 84.71 \\
\hline
\multirow{3}{*}{Fisher} & SVM MCE \cite{TopicID_Fisher_MCE}& 91.9  \\
&MS-CNN~\cite{pappagari2018joint}  & 92.93    \\
& ToBERT & \bf 95.48   \\
& RoBERT &  91.18   \\
\hline

\end{tabular}
\end{normalsize}
}
\caption{Comparison of our results with previous works.}\label{tab:comp_literature}
\end{table}

\section{Conclusions}
In this paper, we presented two methods for long documents using BERT model: RoBERT and ToBERT.
We evaluated our experiments on two classification tasks - customer satisfaction prediction and topic identification - using 3 datasets: CSAT, 20newsgroups and Fisher.
We observed that ToBERT outperforms RoBERT on pre-trained BERT features and fine-tuned BERT features for all our tasks.
Also, we noticed that fine-tuned BERT performs better than pre-trained BERT.
We have shown that both RoBERT and ToBERT improved the simple baselines of taking an average (or the most frequent) of segment-wise predictions for long documents to obtain final prediction.
Position embeddings did not significantly affect our models performance, but slightly improved the accuracy on the CSAT task.
We obtained the best results on Fisher dataset and good improvements for CSAT task compared to the CNN baseline.
It is interesting to note that the longer the average input in a given task, the bigger improvement we observe w.r.t. the baseline for that task.
Our results confirm that both RoBERT and ToBERT can be used for long sequences with competitive performance and quick fine-tuning procedure.
For future work, we shall focus on training models on long documents directly (i.e. in an end-to-end manner).



\bibliographystyle{IEEEbib}
\bibliography{strings,refs}

\end{document}